\documentclass[10pt,twocolumn,letterpaper]{article}

\usepackage[pagenumbers]{cvpr} 

\usepackage{soul}

\definecolor{RefBlue}{rgb}{.5,.5,1}

\usepackage{multirow}
\usepackage{colortbl}
\usepackage{algorithm}
\usepackage{algpseudocode}

\definecolor{cvprblue}{rgb}{0.21,0.49,0.74}
\usepackage[pagebackref,breaklinks,colorlinks,allcolors=cvprblue]{hyperref}

\def\confName{CVPR}
\def\confYear{2026}

\title{The Devil is in Attention Sharing: \\ Improving Complex Non-rigid Image Editing Faithfulness via Attention Synergy}

\author
{
    Zhuo Chen$^{1\star}$ 
    \hspace{2.5mm} Fanyue Wei$^{2\star}$
    \hspace{2.5mm} Runze Xu$^{1}$
    \hspace{2.5mm} Jingjing Li$^{1}$
    \hspace{2.5mm} Lixin Duan$^{1}$
    \hspace{2.5mm} Angela Yao$^{2}$
    \hspace{2.5mm} Wen Li$^{1\dagger}$
    \vspace{1mm}
    \\
    \normalsize{$^{1}$University of Electronic Science and Technology of China}
    \hspace{1mm} \normalsize{$^{2}$National University of Singapore}
    \vspace{1mm}
    \\
    \textbf{Project: \href{https://synps26.github.io/}{\texttt{\textcolor{cyan}{https://synps26.github.io/}}}}
}

\begin{document}


\twocolumn[{
    \renewcommand\twocolumn[1][]{#1}
    \maketitle
    \begin{center}
        \vspace{-15px}
        \includegraphics[width=0.99\linewidth]{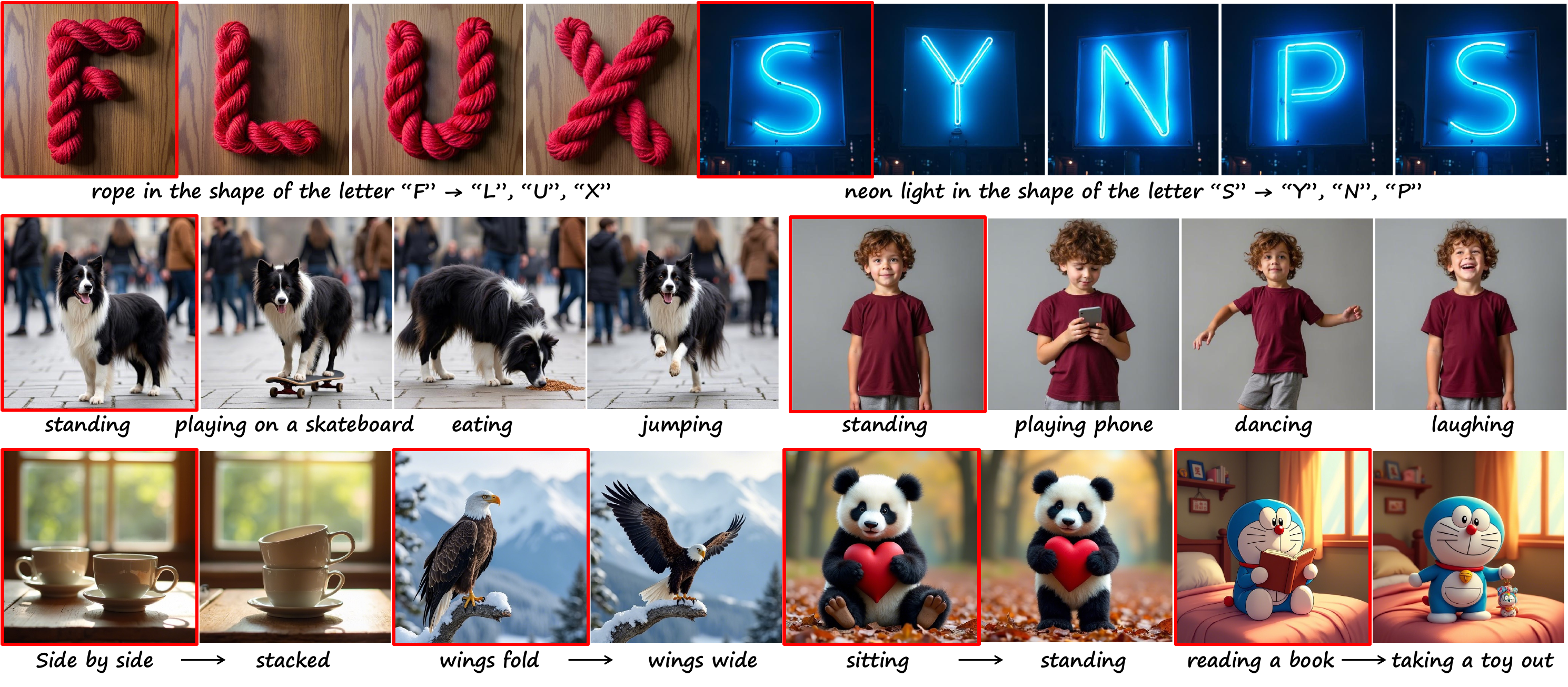}
        \captionsetup{type=figure}
        \vspace{-0.3cm}
        \caption{The editing results produced by our proposed attention synergy mechanism, \textbf{SynPS}. Our method achieves complex training-free non-rigid edits, including challenging tasks such as animal and human pose transformations, image layout transformations, object interactions, and even fine-grained typography editing. The source images are highlighted by red bounding boxes.}
        \label{fig:teaser}
    \end{center}
}]

\footnotetext{$^{\star}$Equal contribution.
\hspace{3mm} $^{\dagger}$Corresponding author.}

\begin{abstract}

Training-free image editing with large diffusion models has become practical, yet faithfully performing complex non-rigid edits (\eg, pose or shape changes) remains highly challenging. We identify a key underlying cause: \emph{attention collapse} in existing attention sharing mechanisms, where either positional embeddings or semantic features dominate visual content retrieval, leading to over-editing or under-editing.
To address this issue, we introduce \textbf{SynPS}, a method that \textbf{Syn}ergistically leverages
\textbf{P}ositional embeddings and \textbf{S}emantic information for faithful non-rigid image editing. We first propose an editing measurement that quantifies the required editing magnitude at each denoising step. Based on this measurement, we design an attention synergy pipeline that dynamically modulates the influence of positional embeddings, enabling SynPS to balance semantic modifications and fidelity preservation.
By adaptively integrating positional and semantic cues, SynPS effectively avoids both over- and under-editing. Extensive experiments on public and newly curated benchmarks demonstrate the superior performance and faithfulness of our approach. 
\end{abstract}    
\vspace{-0.6cm}
\section{Introduction}
\label{sec:intro}

\begin{figure*}[!ht]
    \centering
    \includegraphics[width=\linewidth]{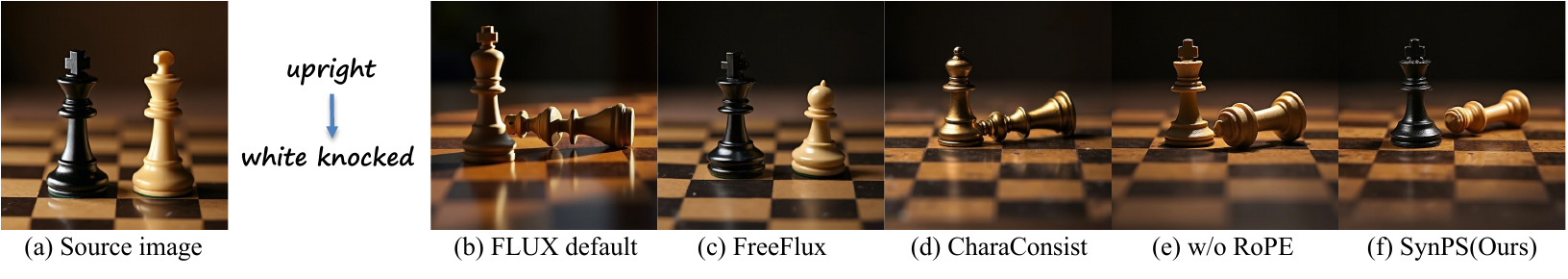}
    \vspace{-0.7cm}
    \caption{Qualitative comparison on the editing instruction ``upright $\rightarrow$ white knocked''. (b) Initialized with the same noise as the source image, the target image is generated using FLUX with default settings conditioned on the target prompt, thereby fully following the textual instruction. (c) FreeFlux produces noticeable \emph{duplicate artifacts} and fails to achieve the intended edit. (d) and (e) show the results of CharaConsist and attention sharing w/o RoPE, respectively. Although the overall structure follows the target prompt, the source color and texture are not well preserved (\ie, \emph{dominated by the prompt}), due to the inaccurate correspondences in CharaConsist and semantic confusion in the \emph{w/o} RoPE setting. (f) Our method better preserves the source appearance while faithfully following the target prompt.}
    \label{fig:compare}
    \vspace{-0.36cm}
\end{figure*}

Image editing~\cite{nichol2021glide,patashnik2021styleclip,hertz2022p2p,tumanyan2023plugandplay,crowson2022vqgan,cao2023masactrl,parmar2023zero} seeks to modify the visual content of an input image according to a given textual instruction, while preserving maximal consistency with the original image. Leveraging recent advances in powerful pre-trained diffusion flow-based generation models~\cite{esser2024sd3, flux2024} such as FLUX~\cite{flux2024}, training-free image editing methods have become practical for real-world applications, eliminating the need for additional fine-tuning or curated datasets~\cite{brooks2023instructpix2pix,labs2025fluxkontext,wu2025qwenimage,cao2025hunyuanimage, wei2024powerful}.

However, editing a source image according to a given textual instruction remains a nontrivial challenge in training-free techniques~\cite{hertz2022p2p, tumanyan2023plugandplay, cao2023masactrl}. Given a pretrained diffusion model, even when initialized with the same noise input, two different textual prompts often yield substantially divergent results (\eg, in human pose or object appearance). This divergence becomes particularly pronounced for complex, non-rigid editing~\cite{kawar2023imagic} instructions (\eg, transforming a \emph{standing} dog into a \emph{sitting} dog), often leading to noticeable artifacts in the edited image. In such cases, either non-target regions or appearance in target region undergo unintended modifications (\eg, background or layout changes), or the target region fails to reflect the desired semantic transformation. Thus, a key challenge lies in faithfully modifying the semantics according to the target prompt while preserving the visual fidelity of the original image.

To mitigate these issues, recent works~\cite{avrahami2025stableflow,wang2024rfsolver,wei2025freeflux,wang2025characonsist} leverage attention sharing mechanisms to better preserve the visual content of the source image. These approaches use tokens from the target image as queries to retrieve relevant visual information from the source image during generation. For instance, CharaConsist~\cite{wang2025characonsist} employs visual features to perform attention sharing, where the source tokens are re-encoded using the position embeddings retrieved from the corresponding target tokens based on their visual correspondence. However, unintended changes often occur when the source and pre-generated target images exhibit layout discrepancies, as shown in Fig.~\ref{fig:compare}~(d). In contrast, FreeFlux~\cite{wei2025freeflux} conducts attention sharing in position-insensitive blocks using target tokens with position embeddings, but this frequently results in source-like images that fail to reflect the intended semantic edits, as shown in Fig.~\ref{fig:compare}~(c).

In this work, we propose \emph{\textbf{SynPS}}, a method that \emph{\textbf{Syn}}ergistically leverages
\emph{\textbf{P}}osition embeddings and \emph{\textbf{S}}emantic information to improve the faithfulness for complex non-rigid image editing. We identify that a key challenge arises from improper query design in attention sharing mechanisms. Specifically, using purely semantic features as queries, as in CharaConsist~\cite{wang2025characonsist}, helps preserve the visual content of the source image but inadvertently loses structural information. Conversely, incorporating position embeddings in queries, as done in prior work~\cite{wei2025freeflux}, restricts the range of attention when retrieving source visual content, often resulting in images that resemble the source rather than the desired edit. We refer to this phenomenon as \emph{attention collapse}, as it is difficult to recover from once it occurs during the image editing process. Therefore, it is crucial to design a pipeline that selectively leverages position embeddings when necessary. The central challenge is determining \emph{when} and \emph{how} to apply them effectively.

To address this challenge, we introduce the SynPS framework. Specifically, to identify when to apply positional embeddings, we first design an \emph{editing measurement} to quantify the magnitude of editing at each step of the denoising process in the diffusion model. We calculate the similarity between the source and target images, as well as the similarity between the source and target textual prompts. The editing measurement is defined as the ratio between intermediate image similarity and text similarity. Intuitively, when this ratio is large, the generated target image might diverge too much from the textual instruction, indicating under-editing, while a small ratio implies over-editing of unfaithful fidelity to the source image. Then, we propose an attention synergy pipeline that dynamically adjusts the effect of position embeddings according to the stepwise editing measurement. When the editing measurement is high, we discard position embeddings to promote diversity in the target image, ensuring that the textual instruction is properly followed. In contrast, when the measurement is low, we incorporate positional embeddings into the queries to reduce unintended changes from the target semantics. When the measurement is in the middle range, we use a scaling weight to balance the effect of position embeddings and semantics in the attention mechanism. We show that the proposed simple yet effective approach significantly mitigates the attention collapse issue, enabling more precise and faithful image editing.

We conduct experiments on the publicly recognized PIE-Bench~\cite{ju2023directinversion} and our newly created diverse benchmarks, with both qualitative and quantitative results demonstrating the effectiveness of our proposed approach. The contributions of this paper are summarized as follows:

\begin{itemize}
    \item We investigate the synergy between position embeddings and semantics in attention sharing for non-rigid image editing.

    \item We propose a training-free strategy \emph{SynPS} to adjust the position embeddings to modulate semantics to resolve identified \emph{attention collapse} issues in non-rigid image editing.

    \item The proposed method achieves new state-of-the-art performance on both PIE-Bench and curated benchmarks.
\end{itemize}

\section{Related Work}
\label{sec:Related Work}

\subsection{Text-to-Image Diffusion Models} 
Diffusion-based models~\cite{ho2020ddpm, rombach2022ldm, podell2023sdxl} have emerged as the dominant paradigm in text-to-image generation and are widely recognized for synthesizing images of exceptional quality. These models adopt the U-Net architecture~\cite{ronneberger2015unet}, which processes visual information through a hierarchy of convolutional and self-attention blocks. Recently, diffusion transformers (DiTs)~\cite{peebles2023dit} mark a major architectural shift. Leveraging transformer scalability and global attention, state-of-the-art models such as SD3~\cite{esser2024sd3} and FLUX~\cite{flux2024} adopt a Multimodal DiT (MM-DiT) that concatenates image and text tokens into a unified sequence, replacing U-Net’s separate attention. Crucially, SD3 applies positional embeddings only at the input layer, whereas FLUX injects Rotary Position Embedding (RoPE)~\cite{su2024rope} into semantic queries and keys at every self-attention layer, resulting in better semantic generation quality. In this paper, we investigate the synergy of positional embeddings and semantics in the attention layers of the RoPE-based MMDiT models like FLUX.

\subsection{Training-Free Non-Rigid Image Editing}
Training-free text-guided image editing methods offer a flexible and efficient way to modify images using natural language. Existing approaches can be broadly categorized into sampling-based and attention-based methods. Sampling-based methods~\cite{song2020ddim, miyake2025npi, mokady2023nti, song2020scorebased, rout2024rfinversion, wang2024rfsolver, deng2024fireflow, kulikov2025flowedit, jiao2025uniedit, liu2022flowsaf, lipman2022flowmgm, wang2024rfsolver, kulikov2025flowedit} manipulate the sampling process by injecting guided noise to achieve more accurate and controllable edits. 
Attention-based methods~\cite{hertz2022p2p, hertz2022p2p, cao2023masactrl, cai2025ditctrl, tumanyan2023plugandplay, alaluf2024crossimageattn, zhu2025kvedit}, in contrast, modify the intermediate attention mechanisms, such as by injecting features or modifying the attention maps to guide semantic changes. 
By sharing attention from source features with target edits to inherit raw appearance and structure, MasaCtrl~\cite{cao2023masactrl} injects source K/V via mutual self-attention to preserve consistency, while DiTCtrl~\cite{cai2025ditctrl} shares attentions within MM-DiT blocks. We further mitigate the attention collapse during sharing for non-rigid image edit.

Most existing training-free editing methods are limited to specific editing types, \eg, object addition, replacement, deletion, or style transfer, which typically preserve the structural feature and spatial layout of the input. In contrast, non-rigid image editing, first introduced by Imagic~\cite{kawar2023imagic}, aims to achieve complex semantic modifications, such as altering object poses or scene layouts, while preserving the overall characteristics and visual identity of the input image. These challenges require a faithful synergy between source and target semantics. To tackle this, StableFlow~\cite{avrahami2025stableflow} empirically identifies vital layers and performs attention sharing only on those layers. Additionally, FreeFlux~\cite{wei2025freeflux} analyzes each block's sensitivity to RoPE and selectively applies attention sharing to those fixed position-insensitive blocks. CharaConsist~\cite{wang2025characonsist} adapts position embeddings based on the correspondences between semantics in the pre-generated target and source images. However, they tend to suffer from \emph{attention collapse} where the generated image either duplicates the source image or is overwhelmed by the target prompt, especially for complex non-rigid image editing, which requires faithful synergy between source and target semantics. In this work, we investigate such issues and further improve non-rigid image editing via attention synergy.

\section{The Attention Collapse Problem}
\label{sec: duplicate issues}

\begin{figure*}[t]
    \centering
    \includegraphics[width=\textwidth]{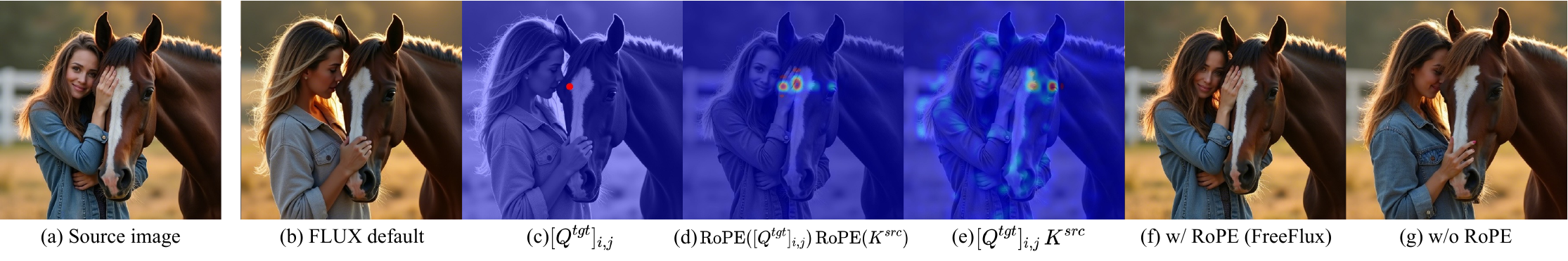}
    \vspace{-0.66cm}
    \caption{Analysis of attention maps \emph{w/} and \emph{w/o} RoPE during attention sharing. (a) The source image is generated from the prompt \texttt{"a woman is hugging a horse."} (b) The target image is generated from \texttt{"a woman is fondling a horse."} using the default FLUX settings. (c) We select a query vector $[Q_{img}]_{i,j}$ at position $(i,j)$ in the target image. (d) and (e) show the attention maps computed between this query vector and the source image, serving as the key. For clarity, we omit the subscript “img” in the figure and use $\text{RoPE}(\cdot)$ instead of $\text{RoPE}(\,\cdot\,,i,j)$ as a simplified notation. In (d), with RoPE injected, attention is localized to spatially adjacent regions. In (e), without RoPE, attention correctly identifies semantically corresponding regions. (f) and (g) show the target images generated using attention sharing \emph{w/} and \emph{w/o} RoPE, respectively.}
    \label{fig:attention_sharing}
    \vspace{-0.36cm}
\end{figure*}

Intuitively, applying attention sharing indiscriminately across all layers and denoising steps inevitably leads to duplication issues dominated by the source image. 
Existing methods~\cite{avrahami2025stableflow, wei2025freeflux} have investigated the use of fixed attention blocks for attention sharing. However, such issues still persist in the edited results due to emphasis on \emph{positional embeddings}. Meanwhile, ~\cite{wang2025characonsist} adjusts position embeddings based on the correspondence between the source image and the pre-generated target image, where the structure is determined by the target prompt. This often leads to cases where the source information is entirely neglected due to the reliance on target \emph{semantics}. This motivates us to investigate the balance between the positional signals of the source and semantics from the target prompt in the attention sharing process. \\
\noindent\textbf{Attention Sharing \emph{w/} and \emph{w/o} RoPE.}
In attention sharing, source image features are gradually injected into the target during the editing process. In detail, target queries attend to a concatenated sequence of target text and source image keys/values as in Eq.~\ref{eq:concat tokens}, enabling target tokens to retrieve and aggregate relevant features from the source image.
\begin{equation}
\begin{aligned}
    \tilde{Q} &= \big[\, Q^{tgt}_{txt} \,;\, \text{RoPE}(Q^{tgt}_{img}) \,\big], \\
    \tilde{K} &= \big[\, K^{tgt}_{txt} \,;\, \text{RoPE}(K^{src}_{img}) \,\big], \\
    \tilde{V} &= \big[\, V^{tgt}_{txt} \,;\, V^{src}_{img} \,\big],
\label{eq:concat tokens}
\end{aligned}
\end{equation}
where RoPE~\cite{su2024rope} embeds the 2D coordinate $(i,j)$ of an image token by rotating its feature vector. Taking the image query as an example, for the query vector $[Q_{img}]_{i,j}$ at position $(i,j)$, the operation is defined as:
\begin{equation} 
\label{eq:rope_op}
    \text{RoPE}([Q_{img}]_{i,j}, i, j) = R_{i,j} [Q_{img}]_{i,j},
\end{equation}
where $R_{i,j}$ is a block-diagonal rotation matrix constructed from the position indices $(i,j)$ whose blocks are 2D rotation matrices. 
The same operation is applied to keys $[K_{img}]_{i,j}$. 
For notational simplicity, we denote the RoPE injection over the full query/key matrices as $\text{RoPE}(Q_{img})$ and $\text{RoPE}(K_{img})$.

For attention sharing without RoPE, simply replace $\text{RoPE}(Q^{tgt}_{img})$ with $Q^{tgt}_{img}$ and $\text{RoPE}(K^{src}_{img})$ with $K^{src}_{img}$. The resulting shared attention is then computed as:
\begin{equation}
\mathrm{Attn}^{tgt} = \mathrm{softmax}\left( \frac{\tilde{Q} (\tilde{K})^T}{\sqrt{d_k}} \right) \tilde{V}.
\label{eq:shared_attention}
\end{equation}
Afterward, we visualize the attention maps in Fig.~\ref{fig:attention_sharing} during the attention sharing process in position-insensitive blocks~\cite{wei2025freeflux}, comparing scenarios \emph{w/} and \emph{w/o} RoPE to further investigate their effects.

\noindent\textbf{Attention Analysis.}
The attention map shown in Fig.~\ref{fig:attention_sharing}~(d) indicates that an attention collapse onto positional embeddings dominates the process, causing query tokens to attend primarily to nearby spatial positions rather than to semantically relevant regions. 
Concretely, query tokens from the region depicting the horse’s eye and white fur are mistakenly attended to adjacent tokens corresponding to the woman’s arm, leading to the duplication artifacts in Fig.~\ref{fig:attention_sharing}~(f). In contrast, when position embeddings are removed, the same query tokens locate semantically relevant tokens across the image, correctly identifying the corresponding horse’s eye and white fur on the other side, as illustrated in \cref{fig:attention_sharing}~(e). As seen in \cref{fig:attention_sharing}~(g), the editing result without position embeddings follows the target prompt while preserving the visual features of the source image.

However, naively removing positional embeddings still yields unsatisfactory results. The query token still exhibits to attend to irrelevant regions (see Fig.~\ref{fig:attention_sharing}~(e)), producing artifacts (\eg, the horse’s fur incorrectly adopts the woman’s yellow hair color in Fig.~\ref{fig:attention_sharing}~(g)). Additionally, as shown in Fig.~\ref{fig:compare}~(e), after directly removing RoPE, the attention collapses onto prompt-dominated semantics, resulting in both chess pieces being rendered in white—similar to the result directly produced by the target prompt in Fig.~\ref{fig:compare}~(b).

Furthermore, given that editing prompts exhibit varying reliance on semantics versus position embeddings, the attention sharing mechanism should be prompt-adaptive rather than fixed throughout the denoising process as in prior works. Motivated by the aforementioned studies, we explore strategies to modulate attention sharing based on the synergy of position embeddings and semantics.

\section{The Attention Synergy Approach}
\label{sec:Methodology}

Building upon the aforementioned analysis of \emph{attention collapse} in Sec.~\ref{sec: duplicate issues}, to tackle the challenge of determining \emph{when} and \emph{how} to apply positional embeddings effectively, we propose the \emph{SynPS} method, which modulates \emph{Syn}ergy between \emph{P}ositional embedding and \emph{S}emantics in Attention for complex non-rigid image editing.
\subsection{Editing Measurement for Attention Synergy}
\label{subsec: Editing Measurement}
To determine \emph{when} to apply positional embeddings during editing, we introduce an editing measurement to quantify the magnitude of editing at each step of the diffusion process. 

Inspired by DeltaEdit~\cite{lyu2023deltaedit}, we compute the cosine similarity between the text tokens of the attention outputs from the source and target branches at each timestep $t$ and for each transformer block $l$. We utilize \emph{Text Similarity} $S_{txt,t}^l$ to quantify the desired degree of editing derived from the prompts, which can be regarded as the semantics guidance in image editing. Additionally, let $\mathrm{Attn}^{l,src}_{txt,t}$ and $\mathrm{Attn}^{l,tgt}_{txt,t}$ denote the attention outputs corresponding to the text tokens for the source and target branches, respectively. The text similarity is calculated as follows:
\begin{equation}
    S_{txt,t}^l = \mathrm{cos\_sim}(\mathrm{Attn}^{l,src}_{txt,t}, \mathrm{Attn}^{l,tgt}_{txt,t}),
\end{equation}
where a larger $S_{txt,t}^l$ indicates a smaller prompt-specified semantic change, which implies that less editing is required at block $l$ and time step $t$.

Similarly, we derive the cosine similarity of the image tokens of the attention outputs from the source and target branches to measure the current editing state. Such \emph{Image Similarity} ($S_{img,t}^l$), reflects how closely the image features of the two branches are aligned at a given timestep $t$ and block $l$. We denote $\mathrm{Attn}^{l,src}_{img,t}$ and $\mathrm{Attn}^{l,tgt}_{img,t}$ as the attention outputs for the image tokens. The image similarity is calculated as:
\begin{equation}
    S_{img,t}^l = \mathrm{cos\_sim}(\mathrm{Attn}^{l,src}_{img,t}, \mathrm{Attn}^{l,tgt}_{img,t}).
\end{equation}
Larger \(S_{img,t}^{\,l}\) indicates stronger alignment of visual features between source and target at block $l$ and timestep $t$.

The \emph{editing measurement} is defined as the ratio of these two similarities. At each timestep $t$, the per-block ratio is derived as $\frac{S_{img,t}^{l}}{S_{txt,t}^{l}}$, then the overall \emph{editing measurement} is the average of these ratios across all $L$ transformer blocks as follows:
\begin{equation}
    M_t = \frac{1}{L} \sum_{l=1}^{L}  \frac{S_{img,t}^{l}}{S_{txt,t}^{l}}.
\end{equation}

The value of $M_t$ provides a clear directive for measuring the faithfulness in the editing process. Intuitively, a large $M_t$ indicates that the edited output diverges from the textual prompt, while a small ratio implies the opposite. 

\begin{algorithm}[ht]
\caption{\emph{SynPS}: Attention Synergy between Positional Embeddings and Semantics}
\label{algo:xxx}
\begin{algorithmic}[1]
\Statex \textbf{Input:} Source prompt $p^{src}$, target prompt $p^{tgt}$, timesteps $T$, FLUX.1-dev backbone $f_\theta$, thresholds $M_{\min}, M_{\max}$, position-insensitive block set $\mathcal{B}_{\text{ins}}$~\cite{wei2025freeflux}
\Statex \textbf{Output:} Edited image $x^{tgt}_0$
\State Initialize $x^{src}_T \sim \mathcal{N}(0, I)$; $x^{tgt}_T \gets x^{src}_T$; $w \gets 1$
\For{$t = T, T-1, \dots, 1$}
    \If{$t < T$}
        \State Set $M_{t+1}$ from previous step statistics \Comment{Mean of per-block ratios at step $t{+}1$}
        \State Set $w \gets 
        \begin{cases}
        0, & M_{t+1} > M_{\max} \\
        1, & M_{t+1} < M_{\min} \\
        \frac{M_{\max} - M_{t+1}}{M_{\max} - M_{\min}}, & \text{otherwise}
        \end{cases}$
    \EndIf
    \For{each transformer block $l$}
        \If{$l \in \mathcal{B}_{\text{ins}}$}
            \State $\tilde{Q} \gets [Q^{tgt}_{txt} \,;\, \text{RoPE}(Q^{tgt}_{img}, w)]$
            \State $\tilde{K} \gets [K^{tgt}_{txt} \,;\, \text{RoPE}(K^{src}_{img}, w)]$
            \State $\tilde{V} \gets [V^{tgt}_{txt} \,;\, V^{src}_{img}]$
            \State $\mathrm{Attn}^{tgt} \gets \mathrm{softmax}\!\left(\frac{\tilde{Q} (\tilde{K})^{\top}}{\sqrt{d_k}}\right) \tilde{V}$ \hfill
        \Else
            \State normal self-attention (no sharing)
        \EndIf
        \State $S_{txt,t}^{l} \gets \mathrm{cos\_sim}(\mathrm{Attn}^{l,src}_{txt,t}, \mathrm{Attn}^{l,tgt}_{txt,t})$
        \State $S_{img,t}^{l} \gets \mathrm{cos\_sim}(\mathrm{Attn}^{l,src}_{img,t}, \mathrm{Attn}^{l,tgt}_{img,t})$
        \State $m_{t}^{l} \gets S_{img,t}^{l} / S_{txt,t}^{l}$
    \EndFor
    \State $M_{t} \gets \frac{1}{L} \sum_{l=1}^{L} m_{t}^{l}$ \Comment{Mean of per-block ratios at $t$}
    \State Denoise one step: $x^{src}_{t-1} \gets f_\theta(x^{src}_{t}, p^{src})$; $x^{tgt}_{t-1} \gets f_\theta(x^{tgt}_{t}, p^{tgt})$
\EndFor
\State \Return $x^{src}_0$, $x^{tgt}_0$
\end{algorithmic}
\end{algorithm}

\subsection{Modulation on Attention Synergy}
\label{subsec:Modulation on Position Embeddings}
On top of the proposed \emph{editing measurement} to determine \emph{when} to apply positional embeddings, we further propose an attention synergy pipeline to deal with \emph{how} to dynamically modulate the effect of positional embeddings on semantics for non-rigid image editing.

The property of RoPE is to encode relative rather than absolute position embeddings of displacement between tokens. The attention score between the query vector $[Q]_{i,j}$ and the key vector $[K]_{i',j'}$ at different positions is thus a function of their relative displacement, as denoted in Eq.~\ref{eq:rope_relative}.
\begin{equation}
    \begin{aligned}
        &\quad\langle \text{RoPE}([Q]_{i,j}, i, j),\; \text{RoPE}([K]_{i',j'}, i', j') \rangle\\
        &= (R_{i,j} [Q]_{i,j})^{\top} (R_{i',j'} [K]_{i',j'}) \\
        &= [Q]_{i,j}^{\top} (R_{i,j}^{\top} R_{i',j'}) [K]_{i',j'} \\
        &= [Q]_{i,j}^{\top} R_{i'-i, j'-j} [K]_{i',j'}. \\
    \end{aligned}
    \label{eq:rope_relative}
\end{equation}

\vspace{-0.02cm}
Since the displacement term is relative, reducing it is equivalent to drawing the tokens closer in the positional space, \ie, reducing the displacement $(i'-i, j'-j)$.
\begin{figure*}[ht]
    \centering
    \includegraphics[width=\textwidth]{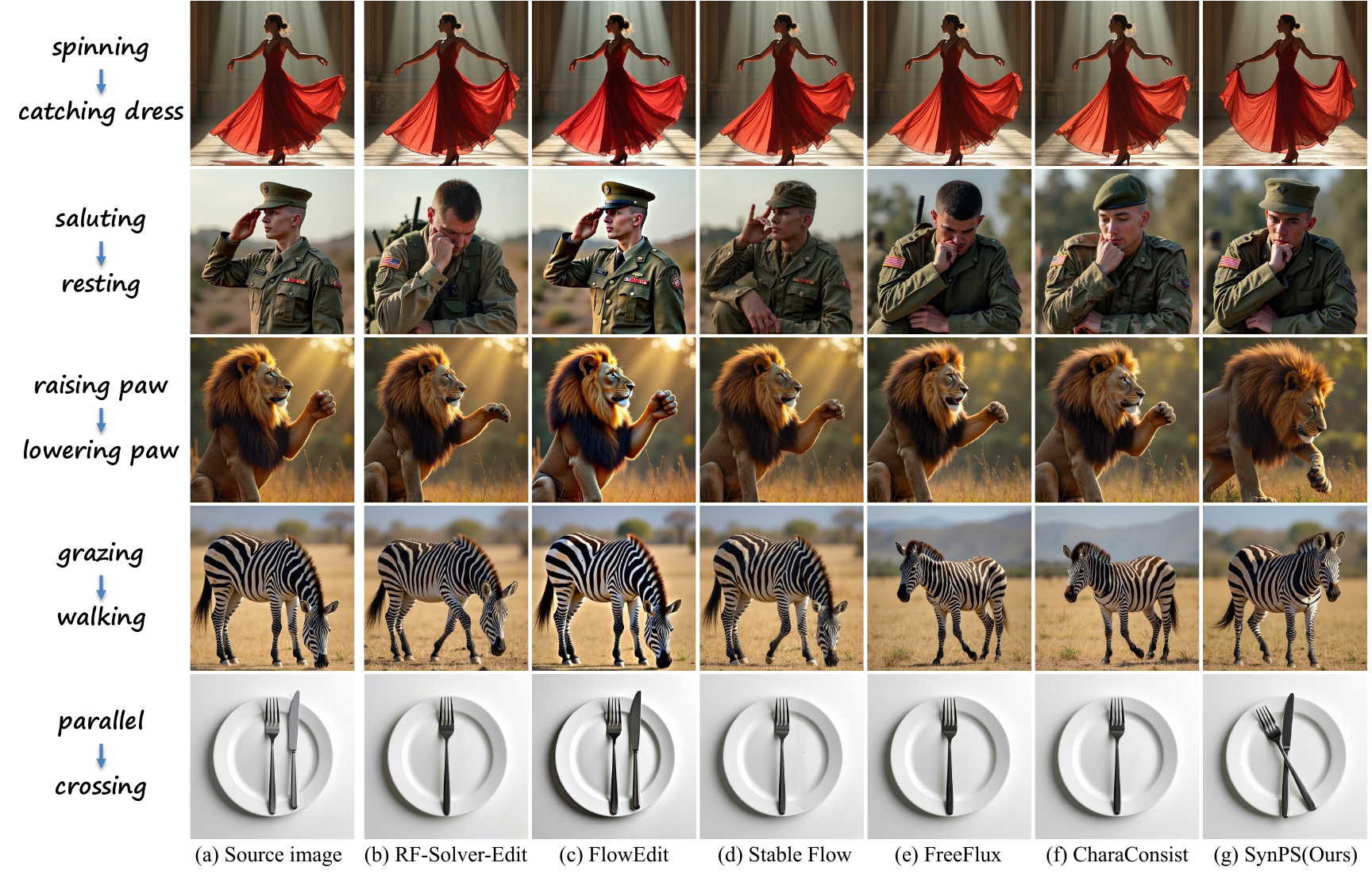}
    \vspace{-0.8cm}
    \caption{Qualitative results on non-rigid editing across benchmarks. As illustrated in the figure, \emph{SynPS}(Ours) achieves the most faithful results compared with all baseline methods.}
    \label{fig:qualitative}
    \vspace{-0.6cm}
\end{figure*}
We further introduce a scaling factor $w \in [0, 1]$ to modulate the effective relative distance on positional embeddings to realize the synergy with semantics in attention sharing.
In detail, such modulation is practically achieved by scaling the position IDs of the query and key tokens with $w$, effectively scaling the rotation angles in the RoPE transformation. When $w=1$, the original positional embeddings are fully preserved to reduce the unintended deformation from target semantics. 
When $w=0$, the positional encoding is effectively discarded, making the attention mechanism position-agnostic. The resulting effect on the attention score between the query vector $[Q]_{i,j}$ and the key vector $[K]_{i',j'}$ is expressed as:
\begin{equation}
\label{eq:rope_modulate}
    \begin{aligned}
        &\quad\langle \text{RoPE}([Q]_{i,j}, w \cdot i, w \cdot j), \text{RoPE}([K]_{i',j'}, w \cdot i', w \cdot j') \rangle \\
        &= [Q]_{i,j}^{\top} R_{w \cdot(i'-i), w \cdot (j'-j)} [K]_{i',j'}.
    \end{aligned}
\end{equation}
By effectively modulating the relative positional relationships encoded by RoPE (further details are included in the supplementary material), this design establishes a continuous spectrum of control between two extremes: from position-aware to completely position-agnostic attention. For notational simplicity, we denote the modulated RoPE injection over the full query and key matrices as $\text{RoPE}(Q, w)$ and $\text{RoPE}(K, w)$.

Based on the analysis of synergy, $M_{t+1}$, we define the adaptive weight $w$ using a piecewise linear function controlled by two thresholds, $M_{\text{max}}$ and $M_{\text{min}}$. The weight at step $t$ is calculated as follows:
\begin{equation}
w =
\begin{cases}
    0, & \text{if } M_{t+1} > M_{\text{max}}, \\
    1, & \text{if } M_{t+1} < M_{\text{min}}, \\
    \frac{M_{\text{max}} - M_{t+1}}{M_{\text{max}} - M_{\text{min}}}, & \text{otherwise.}
\end{cases}
\end{equation}
This formulation ensures that when the image similarity significantly exceeds the desired text similarity ($M_{t+1} > M_{\text{max}}$), indicating under-editing, positional constraints are completely removed ($w=0$) to facilitate greater diversity in the output, ensuring adherence to the textual instruction. Conversely, when the image diverges too much ($M_{t+1} < M_{\text{min}}$), implying over-editing, full positional guidance is enforced ($w=1$) to mitigate unintended deviations from the target semantics. In between these extremes, the weight is smoothly interpolated, providing a fine-grained and adaptive control over the editing process.

The pseudo code of the editing process of the proposed \emph{SynPS} method is presented in Algorithm~\ref{algo:xxx}.

\begin{table*}[ht]
\centering
\renewcommand{\arraystretch}{1.2}
\setlength{\tabcolsep}{6pt}
\resizebox{\textwidth}{!}{%
\begin{tabular}{l||ccccc|ccccc}
\toprule
\multirow{2}{*}{\quad Methods} &
\multicolumn{5}{c}{\textbf{PIE-Bench Change Pose}} &
\multicolumn{5}{|c}{\textbf{Non-Rigid Editing Benchmark}} \\
\cmidrule(lr){2-6}\cmidrule(lr){7-11}
 & GPT-4o$\uparrow$ & GPT-5$\uparrow$ & Gemini-2.5-Pro$\uparrow$ & CLIP$_{img}$ & CLIP$_{txt}\uparrow$ & GPT-4o$\uparrow$ & GPT-5$\uparrow$ & Gemini-2.5-Pro$\uparrow$ & CLIP$_{img}$ & CLIP$_{txt}\uparrow$ \\
\midrule
RF-Solver-Edit & 6.0317 & 4.3250 & 2.9783 & 0.9440 & 0.2664 & 7.0467 & 5.5933 & 4.2400 & 0.9072 & 0.2320 \\
FlowEdit & 4.8217 & 2.8133 & 1.3150 & 0.9731 & 0.2590 & 5.3200 & 3.1100 & 2.7117 & 0.9572 & 0.2260 \\
StableFlow & 4.8083 & 3.4033 & 2.8517 & 0.9712 & 0.2648 & 6.6417 & 5.0183 & 4.2633 & 0.9436 & 0.2287 \\
FreeFlux & 5.5950 & 4.7200 & 3.2417 & 0.9580 & 0.2614 & 7.3900 & 5.9650 & 4.2317 & 0.9180 & 0.2291 \\
Characonsist & 6.3183 & 4.8433 & 3.4933 & 0.9467 & 0.2649 & 7.2683 & 5.5333 & 3.7317 & 0.8899 & 0.2329 \\
\rowcolor{yellow!20} \textbf{SynPS (Ours)} 
& \textbf{6.9900} & \textbf{5.8183} & \textbf{4.1700} & 0.9415 & \textbf{0.2683} 
& \textbf{7.8700} & \textbf{6.6567} & \textbf{5.4250} & 0.9051 & \textbf{0.2344} \\
\bottomrule
\end{tabular}
}%
\vspace{-0.3cm}
\caption{Quantitative comparison with prior methods on non-rigid editing across two benchmarks: PIE-Bench Change Pose and our curated Non-Rigid Editing Benchmark. \textbf{Best results are highlighted in bold.}}
\label{tab:comparison}
\vspace{-0.2cm}
\end{table*}

\section{Experiments}
\label{sec: exps}

\subsection{Experimental Setups}
\noindent\textbf{Implementation Details.}
\label{Implementation Details}
We adopt FLUX.1-dev~\cite{flux2024} as the base text-to-image (T2I) generation model. Unless otherwise stated, we follow the official recommended hyperparameters: 50 sampling steps and a guidance scale of 3.5 by default. We perform attention sharing in position-insensitive blocks following FreeFlux~\cite{wei2025freeflux}, which serves as our \emph{de facto} baseline method.
For the adaptive position embedding weight, we empirically set \(M_{\text{max}}=1\) and \(M_{\text{min}}=0.9\). To mitigate inversion errors across all experiments, we follow previous work~\cite{hertz2022p2p, wei2025freeflux, wang2025characonsist} and use the same random initial noise to generate outputs with both the source prompt and the target prompt. This protocol not only facilitates metric computation but also eliminates the confounding effect of inversion during evaluation, thereby presenting our contribution more clearly.

\begin{table*}[ht]
\centering
\resizebox{\textwidth}{!}{%
\begin{tabular}{l|| l||ccccc}
\toprule
\multirow{2}{*}{\quad \quad \quad Variants} & \multirow{2}{*}{\quad \quad \quad Setting} & \multicolumn{5}{c}{\textbf{Non-Rigid Editing Benchmark}} \\
\cmidrule(lr){3-7}
 &  & GPT-4o$\uparrow$ & GPT-5$\uparrow$ & Gemini-2.5-Pro$\uparrow$ & CLIP$_{img}$ & CLIP$_{txt}\uparrow$ \\
\midrule

Fix Seed FLUX Default & -- & 6.7783 & 5.2383 & 2.3467 & 0.8513 & 0.2364 \\[2pt]
\hline
\multirow{2}{*}{+ Attention Sharing} 
 & w/ RoPE ($w=1.0$) & 7.3900 & 5.9650 & 4.2317 & 0.9180 & 0.2291 \\
 & w/o RoPE ($w=0.0$) & 7.8167 & 6.3750 & 4.7850 & 0.8963 & 0.2366 \\[2pt]
\hline
\multirow{2}{*}{+ SynPS w/ Adaptive $w$} 
 & $M_{\text{min}}=0.8$, $M_{\text{max}}=1.0$ & 7.7483 & 6.3467 & 5.2133 & 0.9052 & 0.2353 \\
 & \cellcolor{yellow!20}$M_{\text{min}}=0.9$, $M_{\text{max}}=1.0$ (Ours)
   & \cellcolor{yellow!20}7.8700
   & \cellcolor{yellow!20}6.6567
   & \cellcolor{yellow!20}5.4250
   & \cellcolor{yellow!20}0.9051
   & \cellcolor{yellow!20}0.2344 \\

\bottomrule
\end{tabular}
}%
\vspace{-0.2cm}
\caption{Ablated results on the Curated Non-rigid Editing Benchmark.}
\label{tab:ablation_nonrigid}
\end{table*}

\noindent\textbf{Comparison Methods.}
\label{Comparison Methods}
We compare our method with state-of-the-art FLUX.1-dev–based~\cite{flux2024} training-free image editing approaches. Among them, RF-Solver-Edit~\cite{wang2024rfsolver}, FlowEdit~\cite{kulikov2025flowedit} and StableFlow~\cite{avrahami2025stableflow} are general-purpose editing methods, while CharaConsist~\cite{wang2025characonsist} and FreeFlux~\cite{wei2025freeflux} are specifically designed for non-rigid editing. All compared baselines are reproduced with their default settings. The implementation details of all comparison methods are provided in the supplementary material.

\noindent\textbf{Benchmarks.}
\label{Benchmarks}
We evaluate on the \emph{ChangePose} subset from PIE-Bench~\cite{ju2023directinversion}, which contains 40 non-rigid editing prompt pairs covering pose, layout, and structural changes. For each prompt pair, we generate results with five random seeds (0--4) to ensure the reliability and stability of the evaluation. To more comprehensively assess effectiveness on non-rigid editing, we further curate the Non-rigid Editing Benchmark, a set of 200 prompt pairs generated by GPT-5~\cite{openai_gpt5_systemcard}, spanning more diverse edit types including pose changes, body-shape variations, facial expression changes, and viewpoint shifts.

\noindent\textbf{Evaluation Metrics.}
\label{Evaluation Metrics}
We report three MLLM-based judgments and two CLIP-based metrics\cite{radford2021clip}.  We employ powerful industry-recognized MLLMs for comprehensive evaluations in non-rigid image editing: GPT-4o~\cite{achiam2023gpt4}, GPT-5~\cite{openai_gpt5_systemcard}, and Gemini-2.5-Pro~\cite{comanici2025gemini}. To reduce stochasticity, we set the temperature to 0 and query each model three times per case, reporting the average score. For CLIP-based metrics, CLIP$_{txt}$ evaluates text–image alignment via similarity between the edited image embedding and the target prompt embedding, while CLIP$_{img}$ assesses source-content preservation via cosine similarity between source and edited image embeddings. Given the complexity of non-rigid editing, trivial duplication of the source image can yield spuriously high CLIP$_{img}$ scores. Consequently, different works~\cite{wei2025freeflux, avrahami2025stableflow, wang2025characonsist} adopt inconsistent treatments of CLIP$_{img}$, which we analyze in detail together with our full evaluation protocols in the supplementary materials.

\subsection{Qualitative Comparison}
As shown in Fig.~\ref{fig:qualitative}, we present qualitative comparisons between our \emph{SynPS} and all compared methods across different challenging scenarios, including pose changes and layout editing. The results show that existing methods either struggle to follow the target prompt to perform intended editing or fail to preserve the appearance of the source image. In contrast, our attention synergy mechanism achieves a more favorable balance between semantic edit fidelity and source image preservation. More visualization cases and analysis are included in the supplementary materials.

\begin{figure*}[ht]
    \centering
    \includegraphics[width=\textwidth]{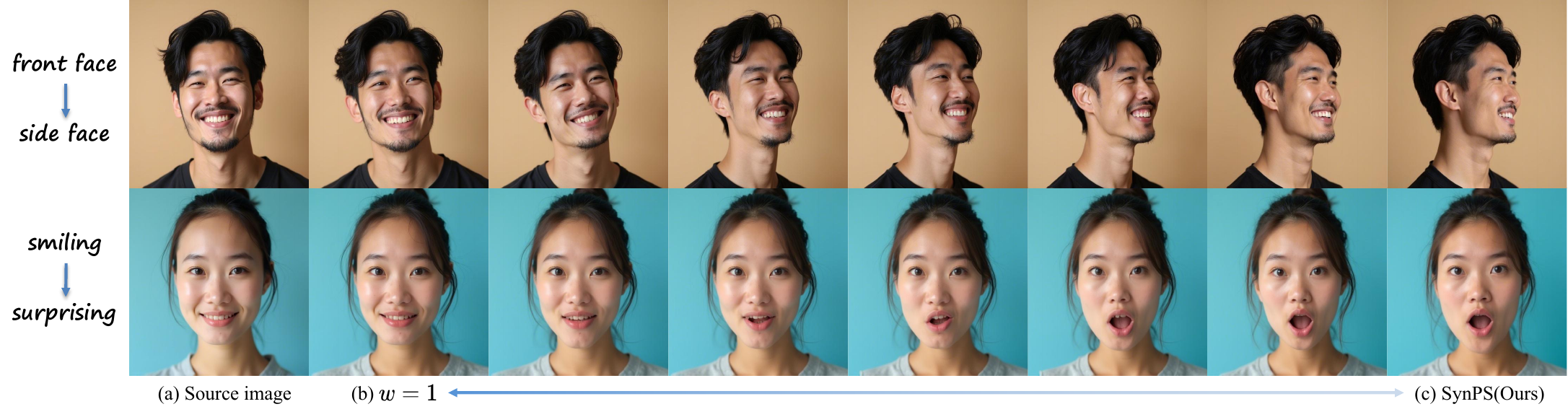}
    \caption{Intermediate results during the interpolation of the attention sharing weight $w$ between 1 and our \emph{SynPS} weights. As $w$ decreases from 1 to our weights, the target image gradually stops replicating the structure of the source image, while still preserving its semantic features and adhering to the prompt guidance.}
    \label{fig:pe_weight_analysis}
\end{figure*}

\subsection{Quantitative Evaluation}
As illustrated in Tab.~\ref{tab:comparison}, across extensive evaluations using three state-of-the-art human-like MLLMs, our method consistently achieves the best overall performance and beats all compared methods by a large margin. For instance, on PIE-Bench, our \emph{SynPS} improves the FreeFlux baseline by $28.6\%$ on Gemini-2.5-Pro score and beats the second-best CharaCosist of $19.3\%$. The quantitative results demonstrate the effectiveness of the proposed \emph{SynPS} to improve complex non-rigid image editing faithfulness via Attention Synergy. 
For CLIP-based metrics, FreeFlux exhibits a strong bias towards the source structure and appearance, often producing duplicate artifacts, which leads to high CLIP$_{img}$ scores but poor CLIP$_{txt}$ performance, as it tends to deviate from the intended edit. 
In contrast, CharaConsist achieves a relatively higher CLIP$_{txt}$ score because its structural guidance is determined by the target prompt, but it yields lower CLIP$_{img}$ as incorrect correspondences frequently lead to unintended modifications. Our method attains a more favorable balance across both metrics, simultaneously preserving desired visual content and adhering to the editing prompt. Further analysis is included in the supplementary materials.

\begin{figure}[t]
    \centering
    \includegraphics[width=\linewidth]{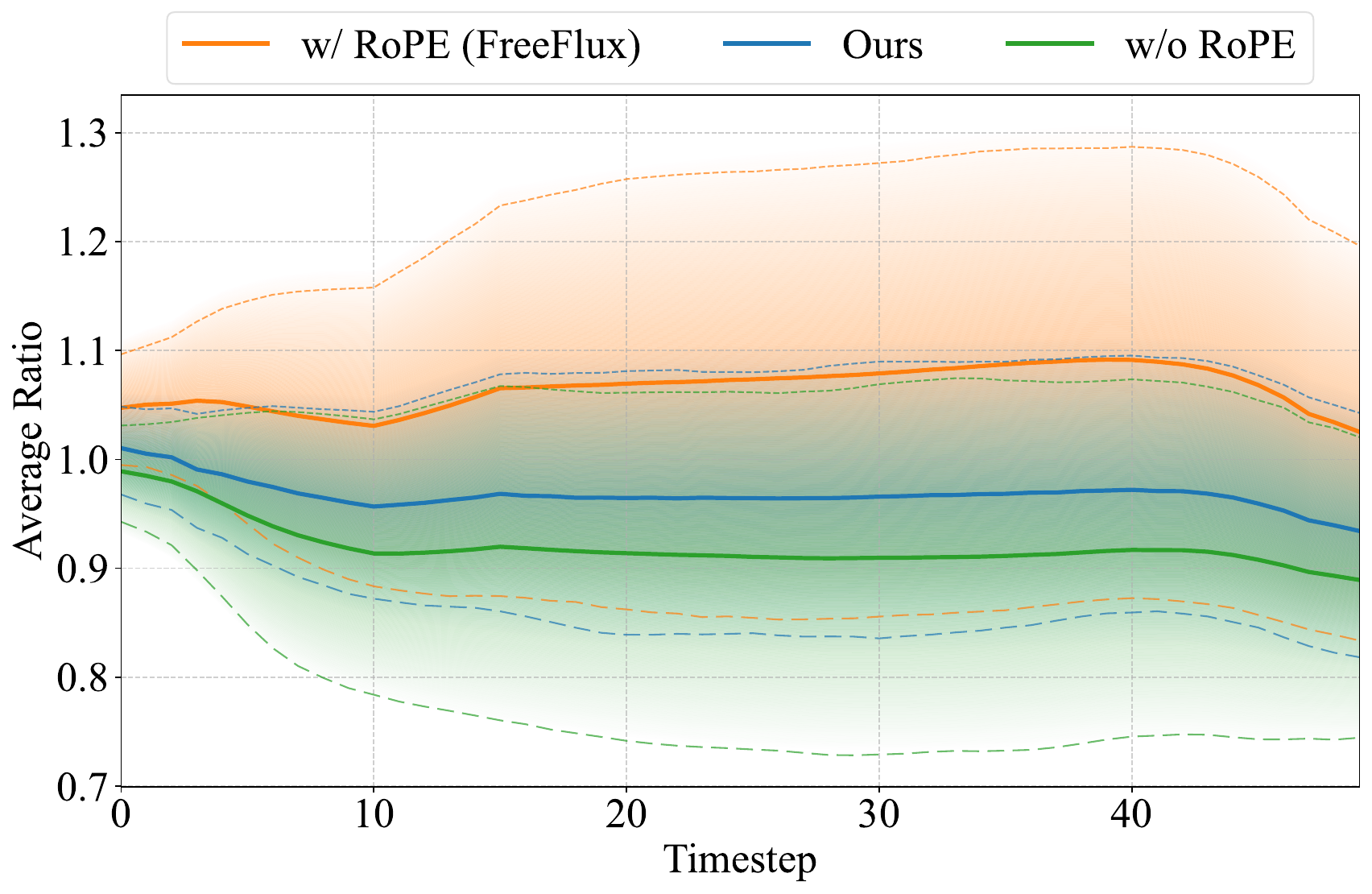}
    \caption{Visualization of the statistics of \emph{editing measurement} over diffusion timesteps for three distinct methods. We evaluate all 200 cases from our curated benchmark. For each method, the solid line represents the mean ratio across all cases. The area between the 20-\emph{th} percentile (long-dashed line) and the 80th percentile (short-dashed line) is highlighted, with the region around the mean ($\pm 1\sigma$) filled with a gradually fading color to indicate the density of the distribution.}
    \label{fig:ratio_timestep}
\end{figure}

\subsection{Analysis on Faithfulness of \emph{\textbf{SynPS}}}
We further validate the effectiveness of the proposed \emph{SynPS} by analyzing the statistics of editing measurement for faithfulness. We aggregate \emph{editing measurement} of \emph{\textbf{SynPS}}, FreeFlux and its variant among all 200 cases on curated benchmarks across the diffusion process. 
As illustrated in Fig.~\ref{fig:ratio_timestep}, \emph{SynPS} stays closest to $1$ across timesteps, indicating faithful editing. Whereas FreeFlux drifts upward, and the ablation \emph{w/o} RoPE falls below $1$.
The spread around the \emph{SynPS} curve is also tighter, suggesting more faithful behavior across cases during the entire editing process. 

As illustrated in Fig.~\ref{fig:pe_weight_analysis}, interpolating the weight between the proposed synergy weight and 1 gradually produces transitions from replicating the source image to preserving its semantic features while adhering to the target prompt's control, exemplified by the face orientation shifting from front-facing (toward the camera) to right-facing and the expression changing from smiling to surprised. This further demonstrates the faithful editability and interpretability of the proposed \emph{SynPS}.

\subsection{Ablation Studies}

We ablate the design choices of our approach on the Non-Rigid Editing Benchmark.
Starting from the baseline ``Fix Seed FLUX Default," we first add Attention Sharing with fixed positional embedding $w=1.0$, which improves Gemini scores by a large margin(from 2.3467 to 4.2317), establishing an effective baseline for comparison.
Additionally, naive removal of position embeddings ($w=0.0$) further boosts to 4.7850, but still suffers from the issue of attention collapse.
We then introduce \emph{SynPS} with $w$ controlled by editing measurement thresholds $(M_{\min}, M_{\max})$. With  $(M_{\min}, M_{\max})=(0.8, 1.0)$ adjusted by intuitive observation from Fig.~\ref{fig:ratio_timestep}, \emph{SynPS} still
achieves promising results (\eg 5.2133 of Gemini score) compared to \emph{Attention Sharing}, demonstrating the robustness of the proposed method. 
Our final configuration $(M_{\min}, M_{\max})=(0.9, 1.0)$ attains the best overall performance across all LLM judges while maintaining strong image-text alignment, validating that tighter thresholds improve instruction faithfulness without sacrificing visual fidelity. More visualization analyses are included in the supplementary materials.

\vspace{-0.2cm}
\section{Conclusion}
This work identifies attention collapse as a key failure mode in training-free image editing under complex non-rigid instructions, stemming from improper reliance on positional embeddings versus semantic features. We introduce \emph{SynPS}, which synergistically couples positional embeddings with semantics by introducing a stepwise editing measurement to quantify edit magnitude along the diffusion trajectory and then dynamically gating positional embeddings to preserve faithfulness. \emph{SynPS} enables non-rigid changes while retaining fidelity. Extensive experiments on public PIE-Bench and curated benchmarks demonstrate its effectiveness.

{
    \small
    \bibliographystyle{ieeenat_fullname}
    \bibliography{main}
}

\clearpage
\maketitlesupplementary

In the supplementary material, we first provide the implementation details of \emph{SynPS} in Sec.~\ref{sec: experimental setups}. Then, we provide the detailed derivation of \emph{SynPS} in Sec.~\ref{sec:details of rope in synps}. Next, we present more visualization cases under complex non-rigid instructions, along with the comparison with the compared baselines in Sec.~\ref{sec:more comparisons}. Additionally, we provide more ablation studies and analysis in Sec.~\ref{sec:more ablation analysis}. 
Finally, we discuss limitations and potential social impact in Sec.~\ref{sec:limitations}.

\section{Experimental Setups}
\label{sec: experimental setups}
\subsection{Implementation Details}
All experiments are conducted at a resolution of 512$\times$512. We follow the FLUX.1-dev official recommended hyperparameters, using 50 sampling steps and a guidance scale of 3.5 by default. In addition, we perform attention sharing in the position-insensitive blocks \texttt{[0, 7, 8, 9, 10, 18, 25, 28, 37, 42, 45, 50, 56]} across all timesteps, following  FreeFlux~\cite{wei2025freeflux}.

\subsection{Details of Compared Methods}
As explained in Sec.~\ref{Comparison Methods}, we compare our method with state-of-the-art training-free image editing baselines under complex non-rigid instructions, adopting the same FLUX.1~\cite{flux2024} as generation backbone. Among them, RF-Solver-Edit~\cite{wang2024rfsolver}, FlowEdit~\cite{kulikov2025flowedit} and StableFlow~\cite{avrahami2025stableflow} are general-purpose editing methods, while CharaConsist~\cite{wang2025characonsist} and FreeFlux~\cite{wei2025freeflux} are specifically designed for non-rigid editing. All compared baselines are reproduced with their default settings. 

For RF-Solver-Edit, StableFlow, CharaConsist, and FreeFlux, we use the same initial noise for all methods and generate the source and target results by applying the source prompt and target prompt, respectively. For RF-Solver-Edit, StableFlow, CharaConsist, and FreeFlux, we use the same initial noise for all methods and generate the source and target results by applying the source prompt and target prompt, respectively. We also follow the official FLUX.1-dev recommended configuration, using 50 sampling steps and a guidance scale of 3.5 by default. This ensures a fair comparison under identical stochastic conditions. Detailed implementations are as follows:
\begin{itemize}
    \item \textbf{RF-Solver-Edit}~\cite{wang2024rfsolver}: 
    We follow the official implementation and set \texttt{inject\_step} to 4, meaning that during the first four denoising steps, the \emph{Value} tokens in blocks [39, 40, \ldots, 56] are replaced from the source to the target.

    \item \textbf{StableFlow}~\cite{avrahami2025stableflow}: 
    We adopt the official implementation, which applies attention sharing with RoPE to the vital blocks [0, 1, 17, 18, 25, 28, 53, 54, 56] across all timesteps.

    \item \textbf{CharaConsist}~\cite{wang2025characonsist}: 
    We adapt the official implementation to fit our evaluation protocol. Following the original design in Characonsist~\cite{wang2025characonsist}, we first perform 11 steps of target pre-generation, then compute the correspondence, and modify the position IDs of the source image accordingly. CharaConsist subsequently conducts point-tracking attention and adaptive token merging from the first sampling step until the 40th step, operating on all single blocks. Unlike our method, which replaces the source-image KV tokens with those of the target image, CharaConsist concatenates the source-image KV tokens to the target-image KV tokens during attention sharing. 
    Additionally, CharaConsist requires carefully engineered prompts consisting of three separate components: foreground, background, and action. Such decomposed prompts are not available in our evaluation setting under complex non-rigid instructions. After extensive analyses, we use the same prompts from our evaluated benchmarks and accordingly bypass the foreground–background mask computation in CharaConsist, applying attention sharing to the entire image without masking. As none of the other compared methods rely on mask computation, this adjustment ensures direct and fair comparison of the core contributions.

    \item \textbf{FreeFlux}~\cite{wei2025freeflux}: 
    We use the official implementation, which performs attention sharing with RoPE in the position-insensitive blocks [0, 7, 8, 9, 10, 18, 25, 28, 37, 42, 45, 50, 56] across all timesteps.

    \item 
    \textbf{FlowEdit}~\cite{kulikov2025flowedit} operates directly on the input image (in contrast to the other methods that modify intermediate attention states under a fixed-seed generative setting), and we provide the FLUX.1-dev generated source image as input to FlowEdit. This places FlowEdit in a fundamentally more challenging setting, rendering the comparison somewhat unfair. Therefore, we emphasize that the comparison with FlowEdit is only intended to analyze the differences between direct generation and inversion-free editing, rather than to demonstrate the superiority of our method. 
     We first generate the source image using the FLUX.1-dev default configuration with the source prompt and feed the generated image into FlowEdit, ensuring that all methods share the same source image. FlowEdit is run with its official recommended settings: 28 inference steps, \texttt{src\_guidance\_scale}=1.5, \texttt{tar\_guidance\_scale}=5.5, \texttt{n\_avg}=1, \texttt{n\_min}=0, \texttt{n\_max}=24, and \texttt{seed}=10.
\end{itemize}


\subsection{Implementation Details of MLLM-based Evaluation}
We employ GPT-4o, GPT-5, and Gemini-2.5-Pro as our evaluation models, which are widely recognized as state-of-the-art MLLMs. All evaluations are conducted using their official APIs. To reduce stochasticity, we set the temperature to 0 and query each model three times per case, reporting the average score.

For each evaluation instance, we provide the source prompt, target prompt, source image, and target image to the MLLM, along with the following instruction prompt:

\begin{quote}
\small
As a Dynamic Transformation Evaluator, your primary role is to assess the quality, realism, and appearance consistency of an object's non-rigid transformation (such as pose, structure, or shape deformation) within a scene. You will be given two images — an original version (source image) and an edited version (target image) — along with the source prompt and target prompt describing the intended transformation.

Your task is to evaluate whether the transformation appears natural, physically plausible, and visually coherent, with special attention to the appearance consistency of the main subject. Specifically, assess:
the realism of the object's pose or structural deformation; the consistency of the subject's appearance, including shape integrity, color tone, texture continuity, lighting conditions, and material properties before and after editing; the preservation of overall scene coherence and non-edited region fidelity; and the accuracy of environmental interactions (e.g., contact points, shadows, reflections, and surface support).

You must provide your evaluation strictly in the following dictionary format: 
\{``score'': 10, ``reason'': ``Explanation here.''\}

Rate the transformation quality on a scale from 0 to 10, where 0 indicates no observable transformation or a visually inconsistent edit, and 10 indicates a perfectly executed, realistic, and appearance-consistent transformation.

When comparing the two images, look for visual evidence of the intended transformation—even subtle changes count. Consider partial success when the transformation partially maintains realism and subject appearance consistency.
\end{quote}

\subsection{CLIP$_{img}$ Analysis}
Non-rigid editing is an inherently complex task that involves multiple aspects, including pose transformation, scene layout changes, object shape deformation, facial expression variation, and viewpoint shift. Such complexity requires a comprehensive evaluation protocol. However, the CLIP$_{img}$ similarity score only measures the global similarity between the source and target images in CLIP latent representation space, and thus cannot evaluate how well the appearance of the transformed subject is preserved. 

Notably, StableFlow~\cite{avrahami2025stableflow} and CharaConsist~\cite{wang2025characonsist} interpret higher CLIP$_{img}$ scores as better, whereas FreeFlux argues the opposite and interprets lower scores as better. In our work, we do not use CLIP$_{img}$ as an editing-quality metric. Instead, we use it solely to assess the similarity to the source image for analyzing whether duplicate artifacts are produced.

\section{Details of RoPE in \emph{SynPS}}
\label{sec:details of rope in synps}
\subsection{Derivation of RoPE}
As shown in Eq.~\ref{eq:rope_op} in the main paper, RoPE~\cite{su2024rope} is applied to the query token
$[Q_{img}]_{i,j}$ at spatial location $(i,j)$:
\begin{equation}
    \text{RoPE}([Q_{img}]_{i,j}, i, j)
    = R_{i,j}\,[Q_{img}]_{i,j},
    \label{eq:rope_op_recap}
\end{equation}
where $R_{i,j}$ is a block-diagonal rotation matrix parameterized by the 2D
position id $[i,j]$.

In FLUX.1-dev~\cite{flux2024}, the position ID of each image token is a
3-dimensional vector $[0,i,j]$.
The Q/K feature dimension is $3072$, split into $24$ attention heads, each
of dimension~$128$.
Each $128$-dimensional head is further partitioned into three contiguous
segments of sizes $[16, 56, 56]$, corresponding to the three position-id
components $0$, $i$, and $j$, respectively.
RoPE is applied to each segment independently using its associated
position-id value.
In the following, we derive the RoPE transformation for a single attention head.

We now focus on the subvector associated with the row index $i$.
Let $[Q_{img}]_{i,j}[16\!:\!72] \in \mathbb{R}^{56}$ denote the 56-dimensional
segment corresponding to the second component of the position id.
The RoPE transformation for this segment can be written in matrix form as
\begin{equation}
    \text{RoPE}([Q_{img}]_{i,j}[16\!:\!72], i, j)
    =
    R_{i}\,[Q_{img}]_{i,j}[16\!:\!72],
    \label{eq:rope_i_main}
\end{equation}
where $R_{i} \in \mathbb{R}^{56 \times 56}$ is the rotation matrix determined
solely by the row index $i$.

Let $q_i \;\triangleq\; [Q_{img}]_{i,j}[16\!:\!72] \in \mathbb{R}^{56}$.
Following RoFormer~\cite{su2024rope}, we decompose $q_i$ into $2$-D subvectors:
\begin{equation}
    \mathbf{q}_i^{(k)}
    =
    \begin{bmatrix}
        q_{i,2k} \\
        q_{i,2k+1}
    \end{bmatrix}
    \in \mathbb{R}^{2},
    \qquad
    \text{for} \quad k = 0,1,\dots,27,
\end{equation}
so that $56 = 2 \times 28$ such subvectors exist.

We assign an angular frequency $\theta_k$ to each pair:
\begin{equation}
    \theta_k
    =
    10000^{-\frac{2k}{d_i}},
    \qquad
    d_i = 56,
    \quad
    \text{for} \quad k = 0,1,\dots,27.
    \label{eq:rope_theta}
\end{equation}
Given the row index $i$, the rotation angle for the $k$-th pair is $\phi_k(i) = i\,\theta_k$. 

\noindent\textbf{2-D rotation.}
RoPE applies a 2-D rotation to each $\mathbf{q}_i^{(k)}$:
\begin{equation}
    \tilde{\mathbf{q}}_i^{(k)}
    =
    R_i^{(k)}
    \mathbf{q}_i^{(k)},
    \qquad
    R_i^{(k)}
    =
    \begin{bmatrix}
        \cos\!\big(\phi_k(i)\big) & -\sin\!\big(\phi_k(i)\big) \\
        \sin\!\big(\phi_k(i)\big) & \phantom{-}\cos\!\big(\phi_k(i)\big)
    \end{bmatrix}.
    \label{eq:rope_Rk}
\end{equation}

Explicitly,
\begin{align}
    \tilde{q}_{i,2k}
    &=
    q_{i,2k}\cos\!\big(\phi_k(i)\big)
    -
    q_{i,2k+1}\sin\!\big(\phi_k(i)\big),
    \\
    \tilde{q}_{i,2k+1}
    &=
    q_{i,2k}\sin\!\big(\phi_k(i)\big)
    +
    q_{i,2k+1}\cos\!\big(\phi_k(i)\big).
\end{align}

\noindent\textbf{Block-diagonal rotation matrix $R_i$.}
Stacking all 28 rotated pairs yields:
\begin{equation}
    \tilde{q}_i
    =
    \big[
        \tilde{q}_{i,0}, \tilde{q}_{i,1}, \dots, \tilde{q}_{i,55}
    \big]^\top
    \in \mathbb{R}^{56}.
\end{equation}
The full rotation matrix is block-diagonal:
\begin{equation}
    R_i
    =
    \mathrm{diag}\!\big(
        R_i^{(0)}, R_i^{(1)}, \dots, R_i^{(27)}
    \big)
    \in \mathbb{R}^{56 \times 56}.
\end{equation}

Thus the RoPE transform on the segment $[16\!:\!72]$ is denoted as:
\begin{equation}
    \text{RoPE}([Q_{img}]_{i,j}[16\!:\!72], i, j)
    =
    R_{i}\,[Q_{img}]_{i,j}[16\!:\!72],
\end{equation}
where $R_i$ is a position-dependent orthogonal linear map determined solely by
the row index $i$.

Applying the same construction to the 16-dimensional segment associated with the
fixed position-id value $0$ and the 56-dimensional segment associated with the
column index $j$ yields three independent rotation blocks.  Together they form
the block-diagonal rotation matrix $R_{i,j}$ for one head.  
Extending this operation to all $24$ attention heads produces the complete RoPE
transformation on the full Q/K feature tensor:
\begin{equation}
    \text{RoPE}([Q_{img}]_{i,j}, i, j)
    = R_{i,j}\,[Q_{img}]_{i,j},
    \label{eq:rope_op_recap}
\end{equation}
where $R_{i,j}$ is the block-diagonal rotation matrix assembled
from the per-head matrices $R_{i,j}$ repeated across all heads, acting on the
entire 3072-dimensional Q/K feature vector.

\subsection{Modulation in \emph{\textbf{SynPS}}}
As shown in Eq.~\ref{eq:rope_modulate} in the main paper, it can be expanded as the 2D image situation:
\begin{equation}
    \begin{aligned}
        &\quad\langle \text{RoPE}([Q]_{i,j}, w \cdot i, w \cdot j),\; \text{RoPE}([K]_{i',j'}, w \cdot i', w \cdot j') \rangle \\
        &= (R_{w i, w j} [Q]_{i,j})^{\top} (R_{w i', w j'} [K]_{i',j'}) \\
        &= [Q]_{i,j}^{\top} (R_{w i, w j}^{\top} R_{w i', w j'}) [K]_{i',j'} \\
        &\overset{(a)}{=} [Q]_{i,j}^{\top} R_{w (i'-i),\; w (j'-j)} [K]_{i',j'},
    \end{aligned}
    \label{eq:rope_relative_scaled}
\end{equation}
where the proof of step~(a) in Eq.~\ref{eq:rope_relative_scaled} is given below.

Recall that for the $k$-th 2D subspace in RoPE, the rotation angle at position
$i$ is given by
\begin{equation}
    \phi_k(i) = i\,\theta_k,
    \label{eq:phi_k_def}
\end{equation}
where $\theta_k$ is the angular frequency associated with that pair of
channels.
When we scale the position index by a factor $w$, i.e., use $w \cdot i$
instead of $i$, the corresponding angle becomes
\begin{equation}
    \tilde{\phi}_k(i)
    \;\triangleq\;
    \phi_k(w i)
    = (w i)\,\theta_k
    = w \,(i\,\theta_k)
    = w\,\phi_k(i).
    \label{eq:phi_k_scaled}
\end{equation}
Thus, scaling the position index by $w$ linearly scales the rotation angle
for every frequency $k$ by the same factor $w$.

For the $k$-th 2D subspace, the rotation matrices at positions $w i$ and
$w i'$ are
\begin{equation}
    \begin{aligned}
        R^{(k)}_{w i}
        &=
        \begin{bmatrix}
            \cos(\tilde{\phi}_k(i)) & -\sin(\tilde{\phi}_k(i)) \\
            \sin(\tilde{\phi}_k(i)) &  \cos(\tilde{\phi}_k(i))
        \end{bmatrix}, \\
        R^{(k)}_{w i'}
        &=
        \begin{bmatrix}
            \cos(\tilde{\phi}_k(i')) & -\sin(\tilde{\phi}_k(i')) \\
            \sin(\tilde{\phi}_k(i')) &  \cos(\tilde{\phi}_k(i'))
        \end{bmatrix}.
    \end{aligned}
\end{equation}
Using the composition rule of planar rotations, we have
\begin{equation}
    \big(R^{(k)}_{w i}\big)^{\top} R^{(k)}_{w i'}
    =
    \begin{bmatrix}
        \cos\!\big(\tilde{\phi}_k(i') - \tilde{\phi}_k(i)\big)
        &
        -\sin\!\big(\tilde{\phi}_k(i') - \tilde{\phi}_k(i)\big)
        \\
        \sin\!\big(\tilde{\phi}_k(i') - \tilde{\phi}_k(i)\big)
        &
        \cos\!\big(\tilde{\phi}_k(i') - \tilde{\phi}_k(i)\big)
    \end{bmatrix}.
\end{equation}
By Eq.~\ref{eq:phi_k_scaled}, the angle difference is
\begin{equation}
    \begin{aligned}
        \tilde{\phi}_k(i') - \tilde{\phi}_k(i)
        &= w\big(\phi_k(i') - \phi_k(i)\big) \\
        &= w(i'-i)\,\theta_k.
    \end{aligned}
\end{equation}
Therefore,
\begin{equation}
    \big(R^{(k)}_{w i}\big)^{\top} R^{(k)}_{w i'}
    =
    R^{(k)}_{w(i'-i)},
\end{equation}
\ie, in the $k$-th subspace, scaling the positions by $w$ results in a
relative rotation whose angle is still proportional to the relative offset
$(i'-i)$, but magnified by a factor of $w$.

Aggregating over all 2D subspaces and extending to the 2D position
$(i,j)$, the same reasoning yields
\begin{equation}
    R_{w i, w j}^{\top} R_{w i', w j'}
    =
    R_{w(i'-i),\,w(j'-j)},
\end{equation}
which leads to the scaled relative-form inner product
\begin{equation}
    \begin{aligned}
        &\quad\langle \text{RoPE}([Q]_{i,j}, w \cdot i, w \cdot j),\; \text{RoPE}([K]_{i',j'}, w \cdot i', w \cdot j') \rangle \\
        &= [Q]_{i,j}^{\top} R_{w (i'-i),\; w (j'-j)} [K]_{i',j'}.
    \end{aligned}
\end{equation}

\begin{table*}[ht]
\centering
\large
\begin{tabular}{l|| l||ccc}
\toprule
\multirow{2}{*}{\quad \quad \quad Variants} & \multirow{2}{*}{\quad \quad \quad Setting} & \multicolumn{3}{c}{\textbf{Non-Rigid Editing Benchmark}} \\
\cmidrule(lr){3-5}
 &  & GPT-5$\uparrow$ & CLIP$_{img}$ & CLIP$_{txt}\uparrow$ \\
\midrule

Fix Seed FLUX Default & -- & 5.2383 & 0.8513 & 0.2364 \\[2pt]
\hline

\multirow{2}{*}{+ Attention Sharing} 
 & w/ RoPE ($w=1.0$) & 5.9650 & 0.9180 & 0.2291 \\
 & w/o RoPE ($w=0.0$) & 6.3750 & 0.8963 & \textbf{0.2366} \\[2pt]
\hline

\multirow{9}{*}{+ SynPS w/ Adaptive $w$} 
 & $M_{\text{min}}=0.7$, $M_{\text{max}}=1.0$ & 6.5572 & 0.9028 & 0.2358 \\
 & $M_{\text{min}}=0.7$, $M_{\text{max}}=1.1$ & 6.5421 & 0.9066 & 0.2355 \\
 & $M_{\text{min}}=0.7$, $M_{\text{max}}=1.2$ & 6.5758 & 0.9087 & 0.2347 \\[2pt]

 & $M_{\text{min}}=0.8$, $M_{\text{max}}=1.0$ & 6.3467 & 0.9052 & 0.2353 \\
 & $M_{\text{min}}=0.8$, $M_{\text{max}}=1.1$ & 6.6566 & 0.9068 & 0.2346 \\
 & $M_{\text{min}}=0.8$, $M_{\text{max}}=1.2$ & 6.5253 & 0.9104 & 0.2343 \\[2pt]

& \cellcolor{yellow!20}$M_{\text{min}}=0.9$, $M_{\text{max}}=1.0$ 
& \cellcolor{yellow!20}\textbf{6.6567}
& \cellcolor{yellow!20}0.9051
& \cellcolor{yellow!20}0.2344 \\
 
 & $M_{\text{min}}=0.9$, $M_{\text{max}}=1.1$ & 6.5633 & 0.9108 & 0.2346 \\
 & $M_{\text{min}}=0.9$, $M_{\text{max}}=1.2$ & 6.3900 & 0.9114 & 0.2336 \\

\bottomrule
\end{tabular}
\caption{Ablated results on the Curated Non-rigid Editing Benchmark.}
\label{tab:ablation_nonrigid}
\end{table*}

\section{More Visulization Comparisions}
\label{sec:more comparisons}

\subsection{Additional Results of SynPS}
\begin{figure}[t]
    \centering
    \includegraphics[width=\linewidth]{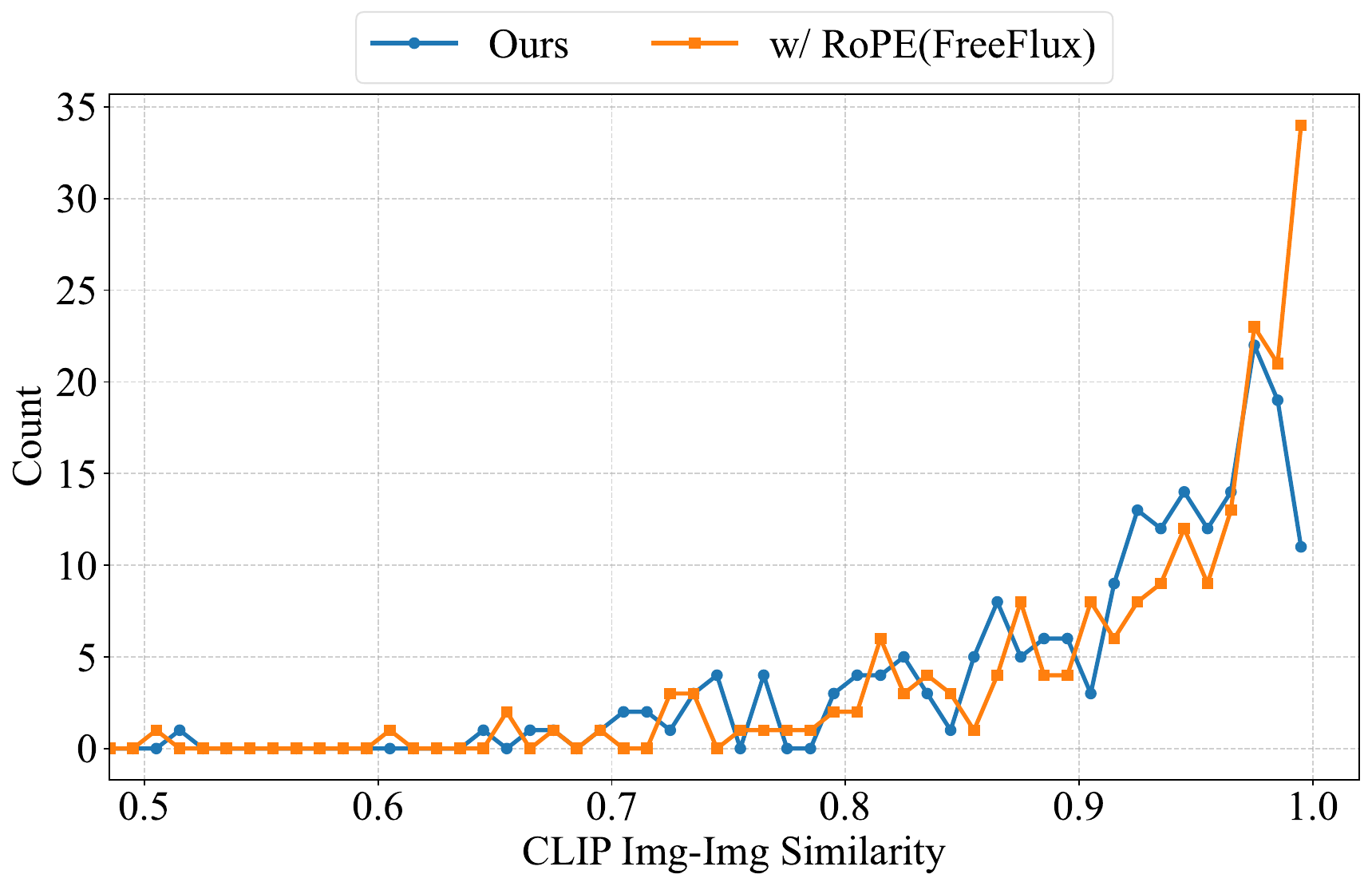}
    \vspace{-0.75cm}
    \caption{Visualization of the distribution of CLIP$_{img}$ similarity scores. We compare the results of the two methods by binning the similarity scores with an interval of 0.01. The horizontal axis represents the CLIP$_{img}$ similarity, while the vertical axis indicates the number of samples falling into each bin.}
    \label{fig:img_img_comparison_line}
    \vspace{-0.6cm}
\end{figure}

\begin{figure*}[t]
    \centering
    \includegraphics[width=0.7\linewidth]{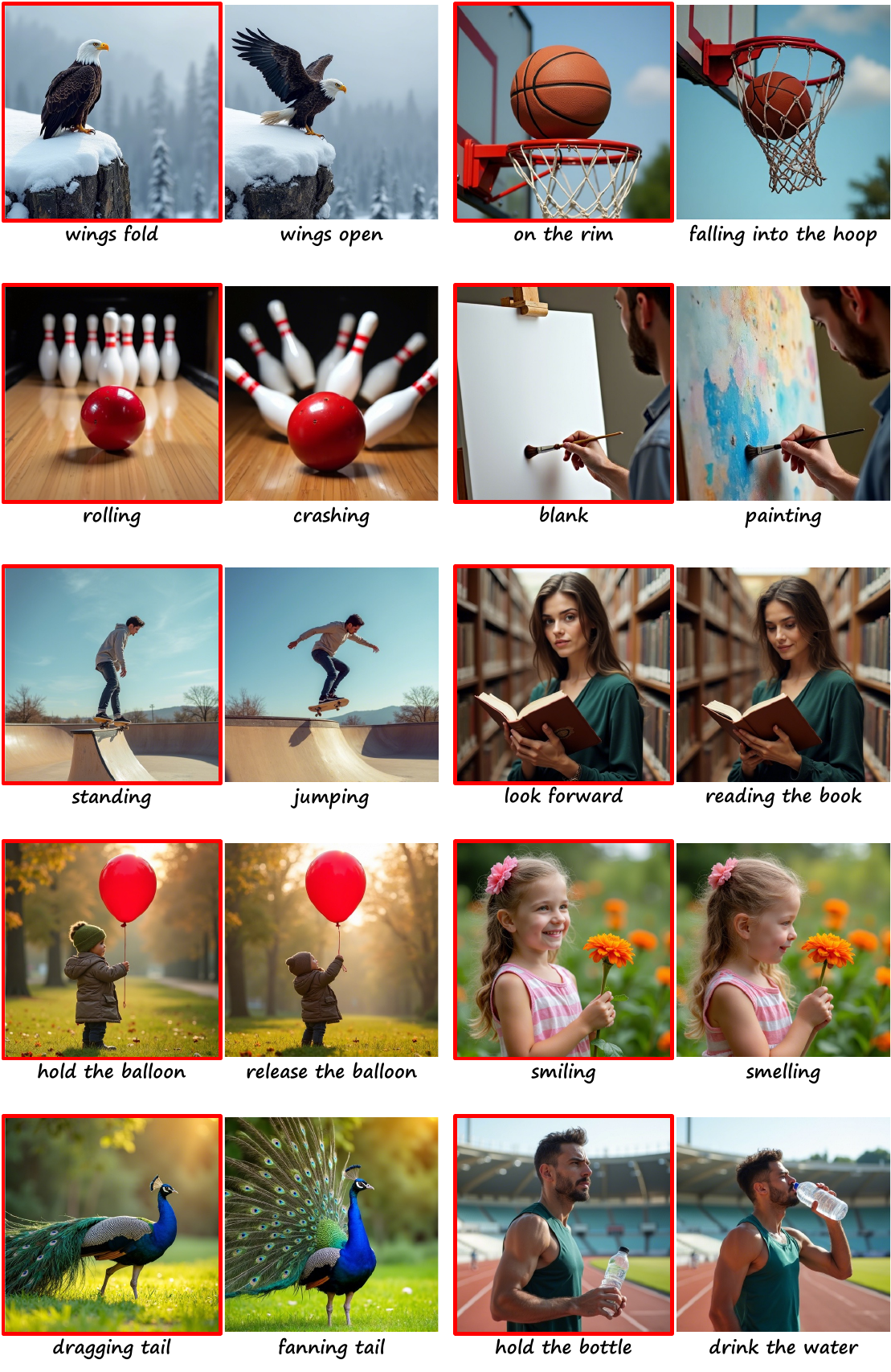}
    \caption{More editing results of \emph{SynPS}.}
    \label{fig:supp_ours}
\end{figure*}

As illustrated in Fig.~\ref{fig:supp_ours}, the proposed \emph{SynPS} is capable of editing the source image with given complex non-rigid prompts.

\subsection{Qualitative Comparison}

\begin{figure*}[t]
    \centering
    \includegraphics[width=\linewidth]{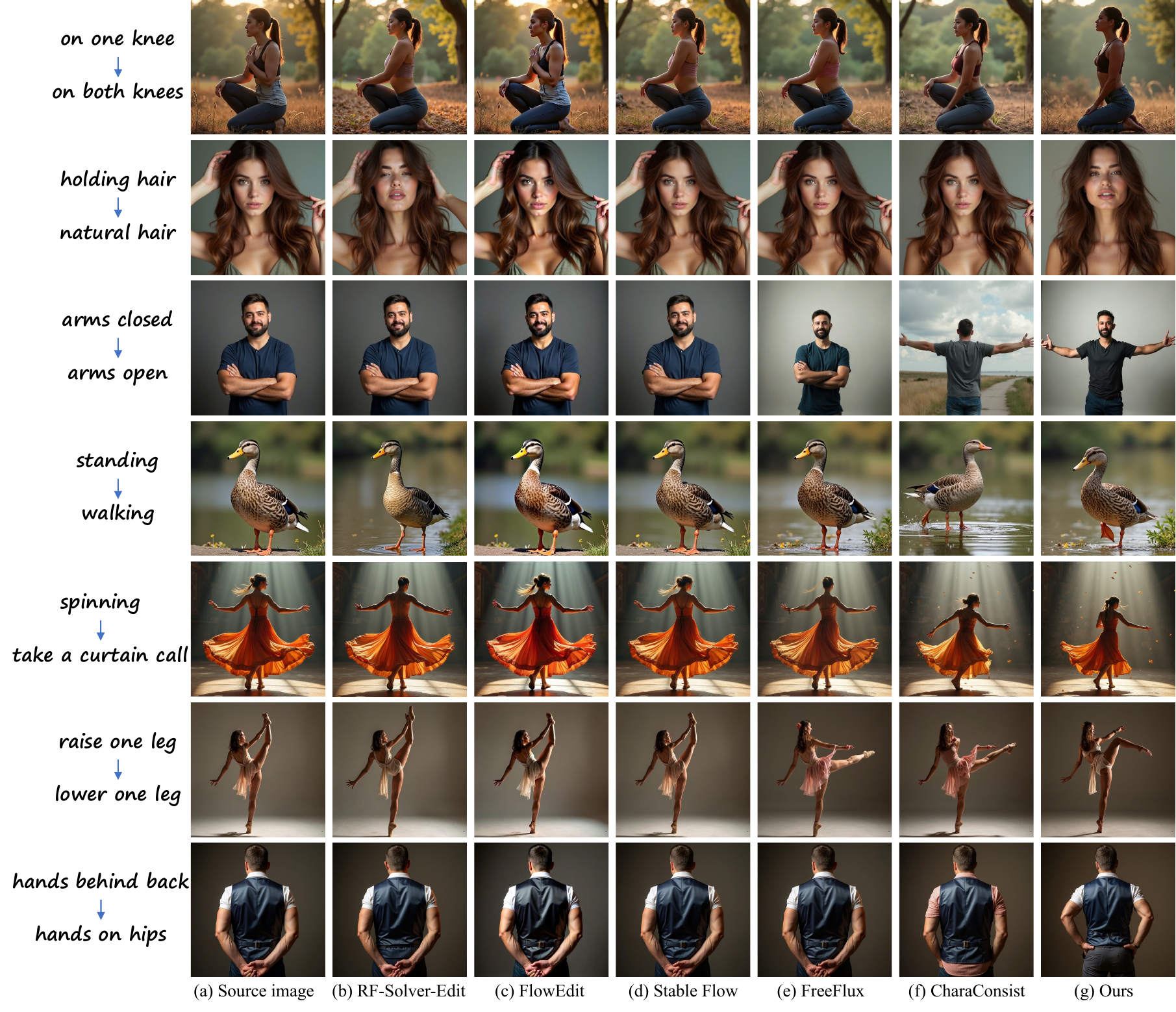}
    \caption{More Qualitative comparisons of compared baselines}
    \label{fig:supp_comparison}
\end{figure*}

As illustrated in Fig.~\ref{fig:supp_comparison}, \emph{Ours} achieves better results than compared baselines.

\section{More Abaltion Analysis}
\label{sec:more ablation analysis}

\subsection{Alleviation of Duplicate Artifacts}

As illustrated in Fig.~\ref{fig:img_img_comparison_line}, We quantitatively define the occurrence of \textit{duplicate artifacts} as instances where the CLIP$_{img}$ similarity between the source and target images exceeds 0.97. By visualizing the distribution of CLIP$_{img}$ similarity scores for both FreeFlux and our method on the benchmark, we observe that FreeFlux yields a substantial number of results with similarity scores surpassing 0.97. Notably, the proportion of samples falling within the range of 0.99-1.0 is significantly higher in FreeFlux compared to our approach. These results further demonstrate that our method effectively mitigates the issue of duplicate generation.

\subsection{Ablations Qualitative Comparison}

\begin{figure*}[t]
    \centering
    \includegraphics[width=\linewidth]{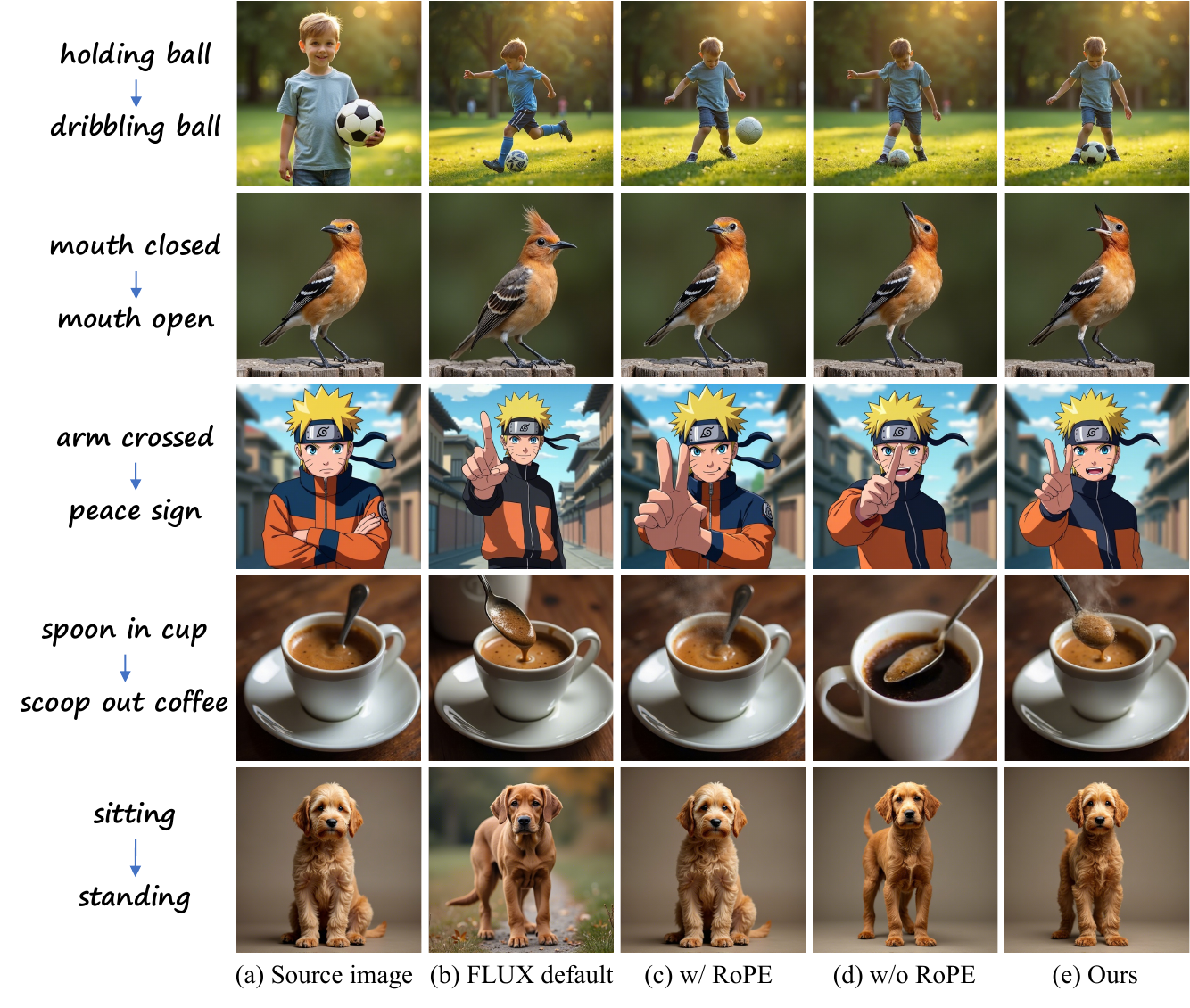}
    \caption{More qualitative ablation studies of \emph{SynPS}.}
    \label{fig:supp_ablation}
\end{figure*}

As illustrated in Fig.~\ref{fig:supp_ablation}, such results demonstrate the effectiveness of the proposed \emph{SynPS}.

\subsection{Hyperparameter Analysis}

As illustrated in Tab.~\ref{tab:ablation_nonrigid}, even with diverse hyperparameters, \emph{SynPS} still achieves promissing results, validating the robustness and effectiveness of the proposed method, especially for the training-free settings.

\section{Discussion}
\label{sec:limitations}

\noindent\textbf{Limitations.}
Our attention synergy mechanism modulates attention sharing by explicitly accounting for the interaction between positional embeddings and semantic information. However, our current design primarily targets non-rigid editing tasks where positional relationships play a crucial role, leaving the exploration of more general editing scenarios to future work. Moreover, when the editing instruction requires structure-preserving transformations—such as color adjustments or style changes—our method becomes less applicable due to the intrinsic characteristics of these tasks, which depend more on appearance-level modifications rather than positional or semantic correspondence. \\
\noindent\textbf{Socail Impact.}
Our methods can modify some fake images with certain instructions, such as human faces or private pets, which may increase the risk of
privacy leakage and portrait forgery. Therefore, users intending to use our technique should apply for authorization to use the respective source images.
Nevertheless, our approach can serve as a tool for AIGC to edit images following the intended instructions.

\end{document}